\documentclass[journal]{IEEEtran}

\usepackage{lineno,hyperref}

\usepackage{fullpage}
\usepackage{blindtext}
\usepackage{amsmath}
\usepackage[pdftex]{graphicx}
\usepackage{amsmath}

\usepackage{algorithmic}
\usepackage{algorithm}

\begin{document}

\title{An Investigation of Interpretability Techniques for Deep Learning in Predictive Process Analytics}

\author{Catarina Moreira,
        Renuka Sindhgatta,
        Chun Ouyang,
        Peter Bruza,
        and Andreas Wichert,\\
        \{catarina.pintomoreira, renuka.sr, c.ouyang, p.bruza\}@qut.edu.au, andreas.wichert@tecnico.ulisboa.pt,
   
\thanks{C. Moreira, R. Sindhgatta, C. Ouyang and P. Bruza are with the School of Information Systems, Science and Engineering Faculty of Queensland University of Technology, 2 George St, Brisbane City QLD 4000, Brisbane, Australia.}%

\thanks{A. Wichert is with the Department of Computer Science and Engineering, Instituto Superior T\'{e}cnico / INESC-ID, University of Lisbon, Av. Prof. Dr. Cavaco Silva, 2744-016 Porto Salvo, Portugal.}
}

\markboth{}
{Moreira \MakeLowercase{\textit{et al.}}}

\maketitle

\begin{abstract}

This paper explores interpretability techniques for two of the most successful learning algorithms in medical decision-making literature: deep neural networks and random forests. We applied these algorithms in a real-world medical dataset containing information about patients with cancer, where we learn models that try to predict the type of cancer of the patient, given their set of medical activity records. 

We explored different algorithms based on neural network architectures using long short term deep neural networks, and random forests. Since there is a growing need to provide decision-makers understandings about the logic of predictions of black boxes, we also explored different techniques that provide interpretations for these classifiers. In one of the techniques, we intercepted some hidden layers of these neural networks and used autoencoders in order to learn what is the representation of the input in the hidden layers. In another, we investigated an interpretable model locally around the random forest's prediction.


Results show learning an interpretable model locally around the model's prediction leads to a higher understanding of why the algorithm is making some decision. Use of local and linear model helps identify the features used in prediction of a specific instance or data point. We see certain distinct features used for predictions that provide useful insights about the type of cancer, along with features that do not generalize well. In addition, the structured deep learning approach using autoencoders provided meaningful prediction insights, which resulted in the identification of nonlinear clusters correspondent to the patients' different types of cancer.
\end{abstract}


\IEEEpeerreviewmaketitle

\section{Introduction}

Deep neural networks are one of the most successful models for prediction in several different fields, specially in medical decision-making~\cite{Esteva19,han2017breast,Beeksma19}. Due to its network representation and activation functions, applications that have in their core deep neural networks can perceive environments, extract and learn different features that characterize the environment, make autonomous decisions and act based on the learnt models~\cite{Goodfellow16}. These network representations together with appropriate regularization techniques allow the incorporation of thousands of hidden layers that contribute to extremely high performances both in classification tasks as well as in forecasts~\cite{LeCun15}.

Although deep learning models achieve very high discriminatory performances, their deep network representation lacks explanatory power. It is hard for decision-makers to understand the logic of predictions of hidden layers inside deep neural networks and obtain insights of why certain decisions were chosen. By explanatory power, we mean models that provide qualitative (and quantitative) inferences about the underlying processes that support their outputs. We have arrived to a stage where high accuracies are no longer sufficient and there is the need to bring together systems that can provide explanatory mechanisms to decision-makers. It has been acknowledged that relying on a couple of evaluation metrics, such as accuracy, precision and recall, is an incomplete description of most real world tasks, and hence new metrics need to be proposed in order to take into account the decision-maker's ability to interpret and understand the predictions of the deep learning algorithms~\cite{doshivelez2017}.

If the goal of deep learning systems is to provide decision-making systems that can assist decision-makers across different fields, including medical decision-making, then one needs systems that have underlying interpretable and explanatory mechanisms that can help decision-makers trust the system's decisions: understand why they work, why they failed, etc~\cite{Miller17}. A misclassification in a patient using a deep learning medical system can have extremely high human costs if one blindly accepts and trusts the system~\cite{Shah18}. This trust can be achieved by creating explanatory models that are able to provide interpretations of why deep learning algorithms are making certain choices~\cite{Kim16}. In this sense, there is a high demand for interpretable deep learning methods that can make the behaviour and predictions of deep learning decision support systems understandable to humans~\cite{Holzinger17}. 

Although opaque decisions are more common in medicine than researchers might be aware of~\cite{Holzinger19}, doctors are constantly confronted with uncertainty, and with data that is incomplete, imbalanced, heterogeneous, noisy, dirty, erroneous, inaccurate and therefore there should be a moral responsibility to provide decision-makers sufficient evidence of why deep learning algorithms are making some predictions in such complicated decision scenarios~\cite{Holzinger19}. Ultimately, medical decisions should belong to the decision-maker rather than the algorithm. The information of the algorithm should therefore complement and augment the knowledge of the decision-maker in scenarios under uncertainty. This leads to a dilemma in terms of the accuracy vs. interpretability tradeoff: either we have models that achieve very high accuracies, such as deep neural networks, but they do not provide any understandings of how the features interact when it comes to predictions; or we have weaker models that provide a reasonable interpretation of the impact different features in the prediction process, such as decision trees, but with much less predictive capacity~\cite{Molnar18}.

\begin{figure}[h!]
    \resizebox{\columnwidth}{!} {
    \includegraphics[scale=0.1]{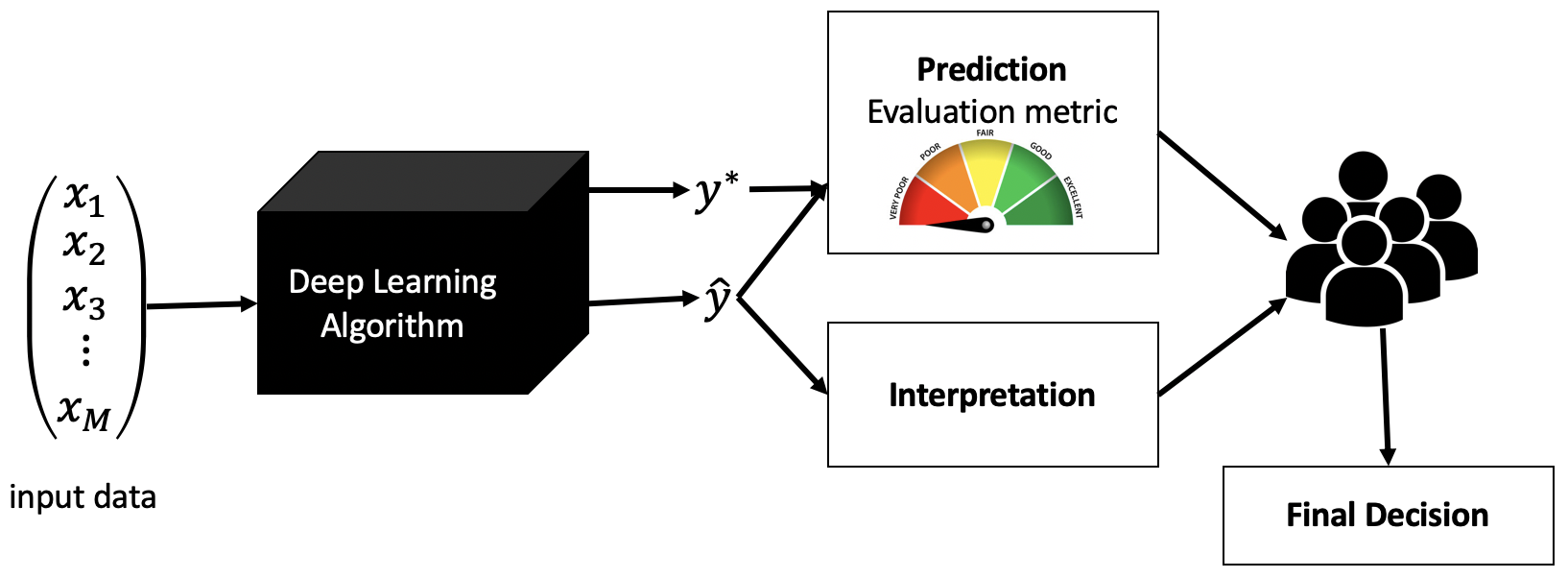}
    }
    \caption{Example of an interpretable deep learning model according to Lipton~\cite{Lipton16}. Given an input vector $x_1, x_2, \dots, x_M$, a deep learning algorithm learns a model by approximating its estimated prediction $\hat{y}$ with the ground truth $y^*$. Ultimately, the decision-maker needs to have an understanding of why the algorithm reached certain predictions.}
    \label{fig:interpretability}
\end{figure}

Figure~\ref{fig:interpretability} illustrates a scenario where a model represented by a black box is learnt by using deep learning algorithms. As suggested by Lipton~\cite{Lipton16}, the learning algorithm should have a performance metric that measures how well the estimator $\hat{y}$ approximates to the ground truth $y^*$ and another metric for interpretability, which measures the degree of understanding of a user towards the explanations provided by the estimated model. Ultimately, the final decision of whether to trust or not the output of the deep learning algorithm resides in the decision-maker. 

In medical decision support systems, predictive tasks using deep learning approaches are hard, due to the fact that doctors are constantly confronted with uncertainty, and with data that is incomplete, imbalanced, heterogeneous, noisy, dirty, erroneous, inaccurate. This data is also expressed in arbitrarily, and unfixed high-dimensional spaces, which makes it hard to model it and to apply machine learning algorithms~\cite{Holzinger14knowledge,Lee16}. Moreover, datasets are small, which makes the learnt models very likely to overfit~\cite{Holzinger16}.  

\begin{figure}[h!]
    \resizebox{\columnwidth}{!} {
    \includegraphics[scale=0.1]{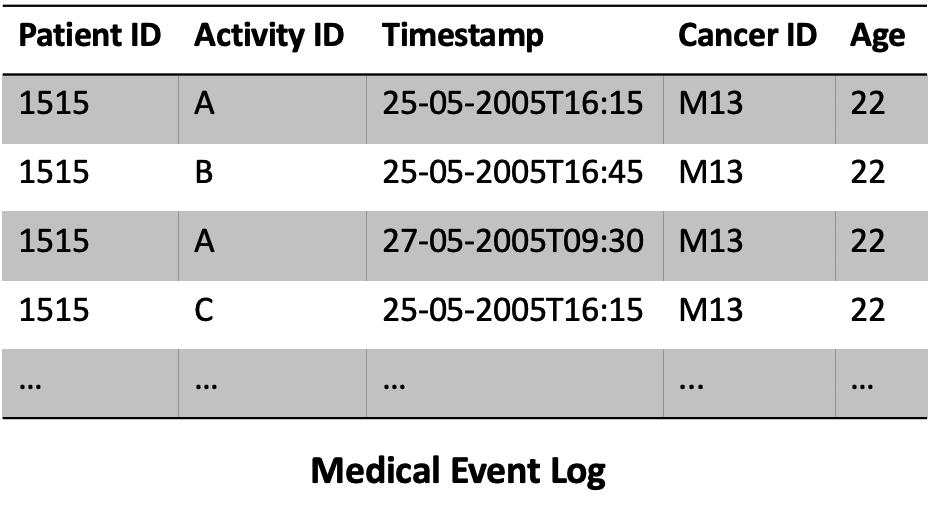}
    }
    \caption{Example of an event log showing different features that are dynamic and change through time and features that are static.}
    \label{fig:event_log}
\end{figure}

In this paper, we investigate explainability mechanisms in deep neural networks and random forests, since these two models are have been successfully applied in different predictive tasks in medical decision-making~\cite{Rahman17,han2017breast}. We explore a real world medical decision event log from the Business Process Intelligence (BPI) Challenge\footnote{\url{https://www.win.tue.nl/bpi/doku.php?id=2011:challenge}} that ran in 2011. This event log corresponds to data that was collected in the Gynecology department from a hospital in the Netherlands. The dataset contains the history of all medical activities undergone by the patient (e.g., blood test, x-rays, medical appointments, etc.), together with information about the treatments and specific information about the patient (e.g., age, number of years spent in treatment, etc). The main challenge with this dataset is that a patient is not defined by an $N$-dimensional feature vector. Instead, a patient is defined by a set of $N$ features that change throughout $T$ timesteps. This means that patients are represented by an unfixed length of medical activities, depending on the severeness of their disease (e.g., a patient might have gone through a set of $70$ different medical activities, together with specific information about other features, and another patient might have gone through a set of 300 medical activities). Figure~\ref{fig:event_log} shows a small example of the event log that we will analyse in this paper.

\begin{figure*}[h!]
    \resizebox{2\columnwidth}{!} {
    \includegraphics[scale=0.1]{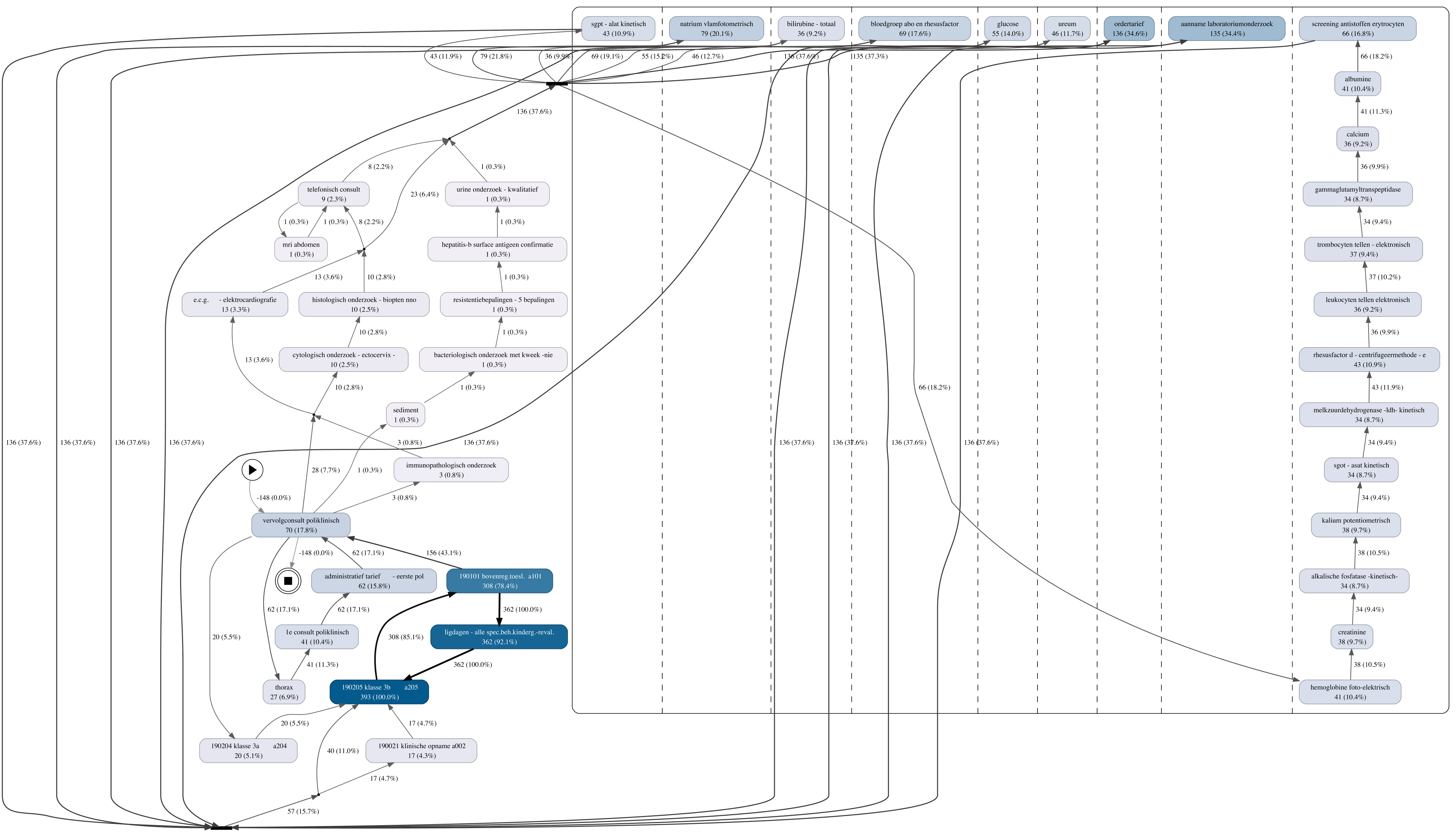}
    }
    \caption{Representing 30\% of the most representative medical activities associated to a single patient diagnosed with cancer of vulva.}
    \label{fig:patient}
\end{figure*}

Given that a patient is represented not by a single $N$-dimensional feature vector, but by a set of $T \times N$ medical activities and features that change throughout time, one can visualise the medical processes that a single patient goes through during treatment. In order to demonstrate the complexity of the medical data that we will cover in this paper, Figure~\ref{fig:patient} shows all medical activities that a single patient who has been diagnosed with cancer of vulva has been through.  

In this sense, we are interested in analysing whether a set of medical activities is targeted to a patient's specific type of cancer. According to Holzinger~\cite{Holzinger14}, health practices should be adjusted to the individual patient and they should be reflected in the hospitals underlying medical decision models.

The two main question addressed by this paper are (1) to understand if patients with specific types of cancer have a targeted and specific set of medical activities associated to them and (2) what type of explainability mechanisms should be involved in this specific task to provide the user (a medical doctors, for instance) the right information that allows the understanding of why the algorithm is making such predictions.

To answer these questions, we use long short term memory (LSTM) neural networks and random forests (RF) to make this prediction. In order to provide explainability mechanisms to the predictions of these models, we explore the usage of autoencoders, where we use these structures to intercept hidden layers of the neural network and try to derive interpretations of clusters that can be found in the data. We also explore a novel explanation technique that explains the predictions of random forests in an interpretable and faithful manner, by learning an interpretable model locally around the prediction~\cite{Ribeiro16}.

In summary, the paper aims to contribute the following:

\begin{enumerate}
\item A deep learning architecture of Long Short Term Neural Networks for Cancer Prediction based on real world event logs of cancer patients

\item Investigate potential interpretations and explanations of the predictions of the Long Short Term Neural Networks using autoencoders.

\item Explore the usage of the LIME framework~\cite{Ribeiro16} in the scope of event logs for medical decision making. LIME consists in a technique that explains the predictions of classifiers in an interpretable manner, by learning an explainable model locally around the prediction.

\end{enumerate}

This paper is organised as follows. In Section~\ref{sec:related_work} we present the main works in the literature that provide interpretable models for black boxes. In Section~\ref{sec:data}, we present the dataset used, how we cleaned it and some initial understandings about the data. In Section~\ref{sec:exp1}, we use deep neural networks to predict the type of cancer of a patient given his track of medical records. In Section~\ref{sec:exp1_interp}, we present an analysis where we use autoencoders to gain deeper insights about how the predictions are being made in the neural network. In Section~\ref{sec:exp2} we model the same data using random forests in order to predict the type of cancer of the patient given the track record of medical activities. In Section~\ref{sec:exp2_interpr}, we apply a local interpretable model-agnostic explanation technique to extract rule-based insights from the predictions of the data. We conclude this paper in Section~\ref{sec:conclusions} where we summarise the main findings in this work.

\section{Related Work: From Predictions to Explanations}\label{sec:related_work}

Over recent years, Deep Learning has demonstrated significant impacts on several predictive tasks in medical decision-making, ranging from advanced decision support systems \cite{mazurowski2008training,esteva2019guide,mantzaris2011genetic}, diagnosis of different types of cancers \cite{lisboa2006use}, Alzheimer's disease \cite{tang2019interpretable,ranasinghe2019neural}, heart disease prediction \cite{ali2019automated}, diabetes diagnosis \cite{liu2019complication}, etc. However, the high performances that these algorithms achieve in terms of accuracy comes at the cost of low explainability and interpretability of the predicted outcomes. Since these classifiers work by computing correlations between features, and since correlation cannot be confused with causation, a solid understanding is required  when making and explaining decisions.

\begin{figure}[h!]
   \resizebox{\columnwidth}{!} {
   \centering
    \includegraphics{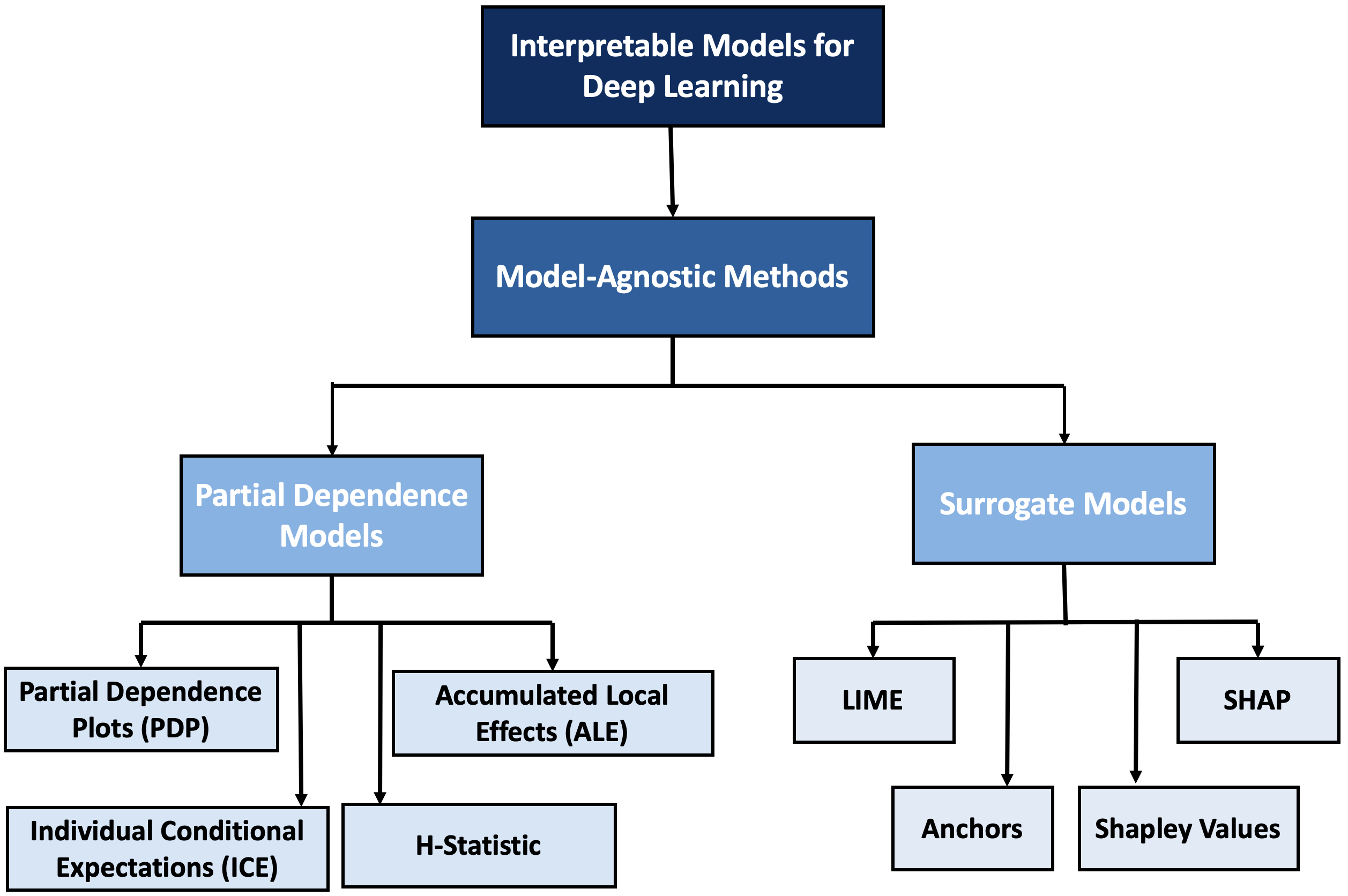}
    }
    \caption{Most relevant model-agnostic methods proposed in the literature.}
    \label{fig:related_work}
\end{figure}

Although explainable models for deep learning are still in their infancy, there are already many works in the literature based on different approaches that provide means to open the "black box"~\cite{Zhang18}. In this work, we focus in model-agnostic methods, which are models that provide explanations and interpretations that can used in any classifier by learning an interpretable model around the classifier's predictions~\cite{Ribeiro16}. The literature in this area can be roughly divided into two main research streams: models based on Partial Dependence Plots (PDP) and Surrogate Models (SM)~\cite{Molnar18}. Figure~\ref{fig:related_work} presents an overview of the most representative model-agnostic methods in the literature.

\subsection{Partial Dependence Plots}

Partial Dependence Plots (PDP) show the marginal effect of at most two features on the predicted outcome of a machine learning model~\cite{Friedman00}. Generally speaking, PDP approaches use Monte Carlo methods to estimate partial functions by calculating averages and marginal effects in the training data. This allows one to get information about how the effect that these averages have in the prediction. In Zhao \& Hastie~\cite{Zhao19}, the authors extended this idea in order to incorporate causal relationships between features and predictions. One main disadvantage of this approach is that it plots the average effect of a feature in the global overall average. The approach also suffers from the independence assumption and assumes that features are not correlated between each other. When computing the marginals, this independence assumption can lead to marginalizations that are not representative of the data.

To address some of the limitations of PDP, two algorithms were proposed: individual conditional expectation and accumulated local effects.

Individual Conditional Expectation (ICE) is a model originally proposed by~\cite{Goldstein15} and is very similar to PDP, but with the difference that it focuses on individual data instances, the individual conditional expectation plots, rather than taking the overall averages. 

Accumulated Local Effects (ALE) Plot was originally proposed by~\cite{Apley16} and differs from PDP by using a small window on the features and making differences between the predictions instead of averages. Since it is not based on comparing averages, ALE is less susceptible to bias and is faster in terms of performance.

There are already packages publicly available with implementations of these algorithms. Some of them are:
\begin{itemize}
    \item R packages that implement PDP methods: $iml$, $pdp$, $DALEX$, $ICEbox$. 
    \item Python libraries that implement PDP methods: $Skater$\footnote{\url{https://github.com/oracle/Skater}}.
\end{itemize}

\subsection{Surrogate Models}

Surrogate models are defined by starting from the input data and the black box model by performing several evaluations of the objective functions with the original model~\cite{Ascione17}. In other words, they are metamodels (or approximation models) that use machine learning methods to approximate the predictions of a black box model, enabling a decision-maker to draw conclusions and interpretations about the black box~\cite{Molnar18}. 

The core idea underlying surrogate models is to use basic interpretable machine learning algorithms, such as linear regression and decision trees, to learn a function using the predictions of the black box model. This means that this regression or decision tree will learn both well classified examples, as well as misclassified ones. Distance functions are used to assess how close the predictions of the surrogate model approximate the blackbox. The general algorithm for surrogate models is presented in Algorithm~\ref{alg:surrogate}~\cite{Molnar18}:

\begin{algorithm} [h!]\label{alg:surrogate}
\caption{General algorithm for surrogate models~\cite{Molnar18}.}
\label{alg:index}
\begin{algorithmic}[1]
\REQUIRE Dataset X used to train black box, Prediction model M\\
\ENSURE  Interpretable Surrogate model I \\
~~\\
\STATE Get the predictions for the selected $X$, using the black box model $M$\\
\STATE Select an interpretable model: linear model, regression tree,... \\
\STATE Train interpretable model on $X$, obtaining model $I$ \\
\STATE Get predictions of interpretable model $I$ for $X$ \\
\STATE Measure the performance of the interpretable model $I$\\

\RETURN Interpretation of $I$

\end{algorithmic}
\end{algorithm}

The most representative model in the literature that applies the surrogate formalism but to explain individual predictions is the Local Interpretable Model-agnostic Explanations (LIME) framework proposed by~\cite{Ribeiro16}, which we will explore in this paper. The main difference between LIME and Algorithm~\ref{alg:surrogate} is that LIME focuses on training local surrogate models to explain individual predictions. This is done by adding a new dataset $X'$ that is a perturbation of the points in dataset $X$. This allows one to see how the features change around these points and how they affect the predictions. The authors explored the Lasso model as the linear interpretation model to approximate the black box. 

An extension of LIME which was also proposed by the same authors is the Anchors model~\cite{Ribeiro18}. The model explains individual predictions by means of easily understandable IF-THEN rules~\cite{Molnar18}, which are called the \textit{Anchors}. To find anchors, the authors use reinforcement techniques to explore the sets of perturbations around the data and how they affect the predictions~\cite{Kaufmann13}.

Another surrogate model that is inspired in game theory is SHAP, SHapley Additive exPlanations, originally proposed by~\cite{Lundberg17}. SHAP is based on Shapley Values, which is a type of game in game theory that focuses in how players distribute payoffs among each other in a fair way~\cite{Shapley53}. In this case, the decision problem is modelled with $n$ features that correspond to the players and the goal is to find a fair way to distribute the weights between each feature. In order to compute this distribution, the authors proposed the KernelSHAP, which estimates for an instance, the contributions of each feature to the prediction. More recently, the authors proposed another kernel, TreeSHAP, that is suitable for tree based machine learning algorithms~\cite{lundberg2018consistent}.

There are already packages publicly available with implementations of these algorithms. Some of them are:
\begin{itemize}
    \item R packages that implement surrogate methods: $iml$, $lime$, $anchorsOnR$,.
    \item Python libraries that implement surrogate methods: $lime$\footnote{\url{https://github.com/marcotcr/lime}}, $skater$, $anchor$\footnote{\url{https://github.com/marcotcr/anchor}}, $shap$\footnote{\url{https://github.com/slundberg/shap}}
\end{itemize}

\section{Dataset Description}\label{sec:data}

The Dutch Academic Hospital Dataset is a publicly dataset made available by the Business Process Intelligence (BPI) challenge in 2011 by a hospital in the Netherlands\footnote{\url{http://www.win.tue.nl/bpi/doku.php?id=2011:challenge}}. The business process intelligence challenge is a competition where organisations make their event logs publicly available, together with specific questions that they would like researchers to address.

The Dutch dataset contains a set of 1142 patients that were diagnosed with a certain type of cancer, together with all the medical activities that they went through in the hospital~\cite{Bose12}. These activities are dynamic and specific to the process of the patient and can describe some specific urine test, in order to try to identify potential tumours in the bladder, tests to the heart, as well as general blood tests and specific cancer-related treatments. The dataset not only contains dynamic features that are connected to the workflow of the process, but it also contains static information, like the patient's age, diagnosis, etc. In total, we have  some 150 291 activities corresponding to all the 1142 patients. Table~\ref{tab:suumary}

\begin{table}[h!]
    \resizebox{\columnwidth}{!} {
    \begin{tabular}{ l | l | c }
        \textbf{Code}     & \textbf{Cancer Name}                          & \textbf{\# Cases}   \\
        \hline
         M11            &  Cancer of Vulva                                          & 60           \\ 
         \textbf{M12}   &  \textbf{Cancer of Vagina  (not representative)}          & \textbf{13}   \\
         M13            &  Cancer of Cervix                                         & 195   \\
         M14            &  Cancer of Corpus Uteri                                   & 95    \\   
         \textbf{M15}   & \textbf{Cancer of Corpus Uteri of type Sarcoma (related to M14)}   & \textbf{11} \\
         M16            & Cancer of the Ovary                                       & 128   \\
         106            & Mix of cancers: cervix, vulva, corpus uteri and vagina    & 113   \\
         \textbf{821}   & \textbf{Cancer of the Ovary (related to M16)}             & \textbf{29}    \\
         \textbf{822}   & \textbf{Cancer of the Cervix (uteri) (related to M13) }   & \textbf{22}   \\
         \textbf{823}   & \textbf{Mix of cancers: corpus uteri, endometrium and ovary} & \textbf{8}  \\
         \textbf{839}   & \textbf{Mix of cancers: ovary, uterine appendages and vulva} & \textbf{14} \\
         \hline
    \end{tabular}
    }
    \caption{Summary of the different types of cancer that can be found in the dataset. Codes 821, 822, 823, 839 and M12 were ignored, since they were not representative in the data.}
    \label{tab:suumary}
\end{table}

The original dataset contains up to 67 features. Many of these features had redundant information. For instance, the diagnosis of the patient was spread across 16 features: {\it Diagnosis, Diagnosis:1, Diagnosis:2, \dots, Diagnosis:15}. This diagnosis attribute can take values such as {\it Squamous cell ca cervix st IIb}, which is a squamous cell carcinoma of the cervix at stage IIb of malignancy. Associated to a diagnosis, the dataset contains a set of 16 features with the diagnosis code: {\it Diagnosis Code, Diagnosis Code:1, Diagnosis Code:2, \dots, Diagnosis Code:15} which can be one of 11 different types of cancer that are specified in Table~\ref{tab:suumary}. The original dataset contains the following attributes:
\begin{itemize}
    \item \textbf{Activity}: describes the medical activities that the patient went through;
    \item \textbf{Department}: identifies the department connected to the activity;
    \item \textbf{Timestamp}: record of the time that the activity took place;
    \item \textbf{Number of executions}: number of times the activity was performed;
    \item \textbf{Activity code}: The dataset does not provide information about this feature;
    \item \textbf{Producer code}: The dataset does not provide information about this feature;
    \item \textbf{Section}: The dataset does not provide information about this feature;
    \item \textbf{Age}: age of the patient;
    \item \textbf{Diagnosis, Diagnosis:1, \dots, Diagnosis:15}: specific diagnosis of the patient, referring to tumours, carcinomas, metasteses, sarcomas, etc;
    \item \textbf{Diagnosis code, Diagnosis code:1, \dots, Diagnosis code:15}: general code specific to a type of cancer;
    \item \textbf{Treatment code, Treatment code:1, \dots, Treatment code:10}: code specific to the treatment applied. The dataset does not provide information about these codes;
    \item \textbf{Diagnosis Treatment Combination ID, Diagnosis Treatment Combination ID:1, \dots, Diagnosis Treatment Combination ID:10}: code specific to the combination of the treatment and the diagnosis of the patient. The dataset does not provide information about these codes;
    \item \textbf{Start Date, Start Date:1, \dots, Start Date:15}: start date of the activity of the patient;
    \item \textbf{End Date, End Date:1, \dots, End Date:15}: end date of the activity of the patient;
    \item \textbf{Specialism code, Specialism code:1, Specialism code:2}: code specific to the specialism of the diagnosis of the patient. The dataset does not provide information about these codes;
\end{itemize}

The data cleaning process was conducted in the following steps:
\begin{itemize}
    \item {\bf Missing values.} The dataset contained 455 instances of patients who did not have any diagnosis code. The diagnosis code was spread across 16 different features (\textit{Diagnosis, Diagnosis:1, \dots, Diagnosis:15}). In many cases, missing values were found in the remaining 15 features. For the cases where this information was not available across other features, we were able to infer the type of diagnosis based on patients who shared similar activities and treatment codes.
    
    \item {\bf Time features.} The dataset contains 33 time related features: tart Date, Start Date:1, \dots, End Date:15, End Date, End Date:1, \dots, End Date:15 and Timestamp. The start and end dates had a huge amount of missing information and it was difficult to make any inferences about the distribution of the missing data. For that reason, we ignored these features, and instead, we created a new feature {\it years} that corresponds to the total amount of years a patient was under treatment. This information was taken by making the different between the timestamp recorded for the first and last activities.
    
    \item {\bf Repeated features.} Features whose information was spread around multiple features (e.g. Treatment code, Treatment code:1, \dots, Treatment code:10) were collapsed into a single feature representing the last event recorded.
\end{itemize}

\begin{figure}[h!]
	 \resizebox{\columnwidth}{!} {
	 \includegraphics{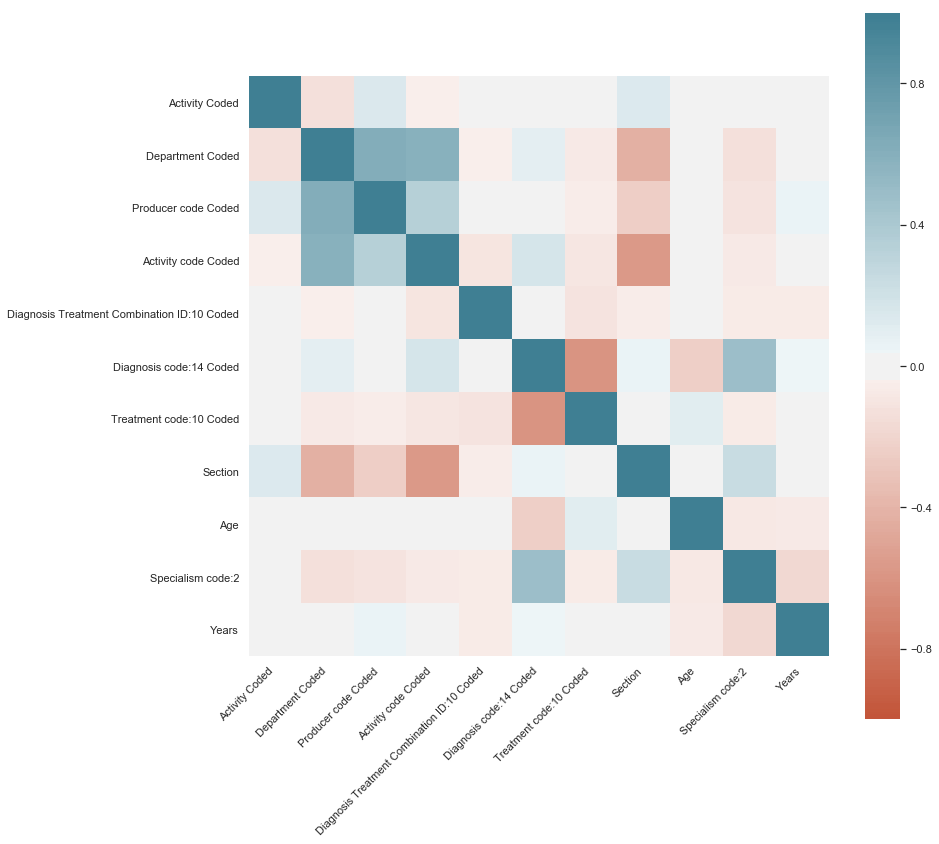}
	 }
	  \caption{Correlation between features in the medical event log, after balancing and cleaning the data.}
	   \label{fig:correlation}	
\end{figure}

After cleaning the dataset, we ended up with 12 features: Activity, Department, Number of executions, Activity code, Producer code, Section, Age, Diagnosis Code, Treatment code, Diagnosis Treatment Combination ID and years. We analysed the distribution and correlation of the features of the dataset. Figure~\ref{fig:correlation} show these relationships. 

An initial look at the correlation map of the feature shows that the features do not show many correlations with the diagnosis code. This preliminary analysis suggested that there can be a template set of procedures to apply to patients that show some potential symptoms of cancer, however it does not seem to be any targeted set of procedures that a patient goes through that is specific to a type of cancer. This lack of correlation can already indicate that machine learning approached might not have very high accuracies in this specific dataset for the task of cancer prediction.

\section{Experiment I: Explanatory Mechanisms for Predictions Using Deep Neural Networks}\label{sec:exp1}

In this section, we test the hypothesis that, in theory, patients with a specific type of cancer should be associated to a more targeted set of medical activities. We test this hypothesis by formulating our problem under a deep neural network approach.

\subsection{Problem Definition}\label{sec:exp1_problem}

Contrary to traditional deep learning approaches in the literature, where a patient is defined by a single $F$-dimensional feature vector, when using event logs, we have a description of daily (or even by hour / minutes / seconds) medical activities associated to a patient. This means that a single patient $X^(i)$ from a set of $M$ patients, $X^(i) \in \{X^(1), X^(2), \dots, X^(M)\}$, is defined by a set of $F$ features that are both dynamic (showing the evolution of medical activities throughout time) and static (features concerned with the number of years the patient stays in the hospital). The length, $T$, of these features is also dynamic, meaning that a patient that stays 2 years in the hospital can have records of more than 1000 medical activities associated to him, while another patient that spends 1 month in the hospital can only have 20 activities in his records, for instance. Therefore, a set of patients is represented by a tensor with dimensions $(M \times L \times F)$ where $M$ corresponds to the total number of patients, $L$ corresponds to the length of the patient's medical records and $D$ is the set of features associated with the patients. Each patient is also associated to a label that corresponds to the specific type of cancer that he has been diagnosed with, $Y^{(1)}$, where  $Y^(i) \in \{Y^(1), Y^(2), \dots, Y^(M)\}$. This is the class that we are interested in predicting.

\[
X^{(1)} = 
\begin{pmatrix} 
f_{1,1}^{(1)}       & f_{1,2}^{(1)}     & \hdots    & f_{1,F}^{(1)}    \\
f_{2,1}^{(1)}       & f_{2,2}^{(1)}     & \hdots    & f_{2,F}^{(1)}   \\
\vdots              & \vdots            & \ddots    & \vdots            \\
f_{T,1}^{(1)}       & f_{T,2}^{(1)}     & \hdots    & f_{1,F}^{(1)})   \\
\end{pmatrix}_{T \times F}
\hdots ~~~~~~~~~~~~~~~~~~
\]
\[
X^{(M)} = 
\begin{pmatrix} 
f_{1,1}^{(M)}    & f_{1,2}^{(1)}       & \hdots    & f_{1,F}^{(M)}    \\
f_{2,1}^{(M)}    & f_{2,2}^{(1)}        & \hdots    & f_{2,F}^{(M)}   \\
\vdots           \vdots             & \ddots    & \vdots            \\
f_{T,1}^{(M)}    & f_{T,2}^{(1)}        & \hdots    & f_{1,F}^{(M)})   \\
\end{pmatrix}_{T \times F}  ~~~~~~~~~~~~~~~~~~~~~~~
\]
\[
Y = 
\begin{pmatrix} 
class^{(1)}     \\
class^{(2)}    \\
\vdots          \\
class^{(M)}     \\
\end{pmatrix}_{M \times 1 }  ~~~~~~~~~~~~~~~~~~~~~~~~~~~~~~~~ 
\]

\subsection{Exploring Deep Learning Architectures for Cancer Prediction}\label{sec:exp1_algo}

In the scope of this work, we analyse a trail of medical activities and appointments associated to a patient. This set of medical activities is recorded in a given order, which suggests dependence between them. 

Since the nature of the data analysed in this work is dynamic, one needs a supervised learning mechanism that is able to cope with data that has a strong and meaningful dependency between features and that is also able to keep in memory all the information from previous time steps. For these reasons, we opted for a Recurrent Neural network (RNN). RNNs were originally proposed by~\cite{Williams1989rnn} and consist in a neural network with hidden units capable of analysing streams of data and that has reveled to be effective in many different applications which require a dependency in previous computations during the learning process, such as text classification~\cite{Liu06}, speech~\cite{Graves13}, or even DNA sequences~\cite{Shen18}. One important characteristic of RNNs is that they share the same weights across all training steps, which is something that does not occur in traditional deep neural network models.

In this work, we explored two different types of Recurrent Neural Networks:

\begin{itemize}
    
\item {\bf Long Short Term Memory (LSTM) Neural Networks:} are a type of recurrent neural networks that are particularly suitable  for applications where there are very long time lags of unknown sizes between important events. They provide a solution for the vanishing and exploding gradient problems by using memory cells~\cite{Hochreiter97}. These memory cells, $C_t$ are composed of a self recurrent neuron together with three gates: an input gate, $i_t$, an output gate, $o_t$, and a forget gate, $f_t$. These gates are used to regulate the amount of information that goes in / out of the cell. Information on a new input will be accumulated to the memory cell if $i_t$ is activated. Additionally, the past memory cell status, $C_{t-1}$ can be {\it forgotten} if $f_t$ is activated. The information on $C_t$ will be propagated to $h_t$ based on the activation of output gate $o_t$. Based in the activation functions, new candidates for the memory cell, $\tilde{C}$, are created. \\

\item {\bf Bidirectional Long Short Term Neural Networks (BiLSTM):} are also a type of recurrent neural network that connect two hidden layers of opposite directions to the same output, which was originally proposed by~\cite{Schuster97birnn}. The motivation of bidirectional neural networks is due to certain contexts specific to datasets. It is not enough to learn from the past to predict the future activities, but also it should be possible to look at the future activities in order to fix the current predictions. 


\end{itemize}


\subsection{Predicting Patient's Type of Cancer}\label{sec:exp1_pred}

\begin{figure*}[h!]
    \centering
    \includegraphics[scale=0.13]{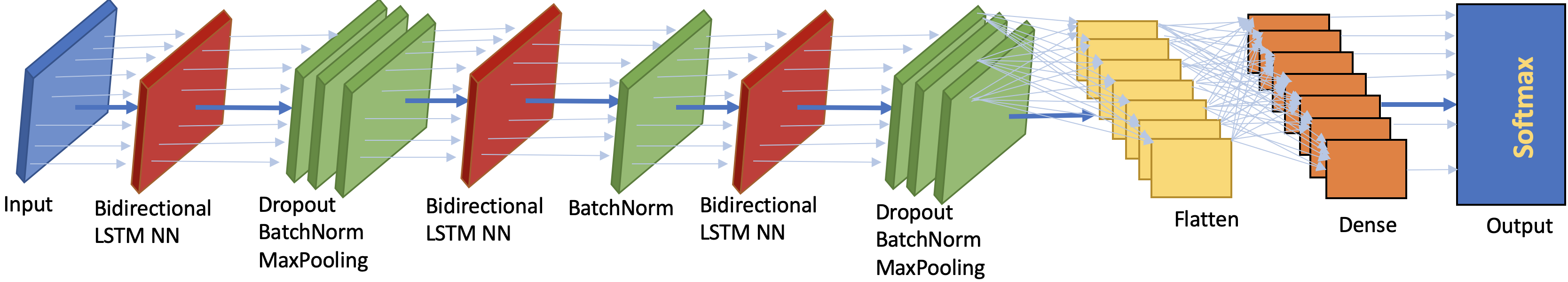}
    \caption{Deep neural network architecture used in our experimental setup.}
    \label{fig:deep_arch}
\end{figure*}

In this section, we test the hypothesis that, in theory, patients with a specific type of cancer should be associated to a more targeted set of medical activities. To validate this, we performed a cross validation setting with a train/test set split of 80\% / 20\% over the network architecture illustrated in Figure~\ref{fig:deep_arch}. Table~\ref{tab:results_exp1.1} illustrates the results obtained.



\begin{table}[h!]
    \resizebox{\columnwidth}{!} {
    \begin{tabular}{l | c | c | c | c |}    
        & {\bf Nodes}   & {\bf Epochs}    & {\bf Accuracy}    & {\bf Loss}  \\
        \hline
{\bf Deep NN}           & 25       & 30       & 0.468   & 1.297       \\
{\bf LSTM NN}           & \textbf{20}  & \textbf{200} . & \textbf{0.552}  & \textbf{1.216}   \\
           {\bf BiLSTM NN}  & 20       & 150       & 0.517         & 1.230    \\
           \hline
    \end{tabular}
   }
    \caption{Results obtained after conducting a cross validation grid search method over the distribution of neurons and epochs using the architecture illustrated in Figure~\ref{fig:deep_arch}. Best results were found when using a deep Long Short Term Memory recurrent neural network during 200 epochs and 20 neurons in the hidden layers.}
    \label{tab:results_exp1.1}
\end{table}

One major challenge with deep neural networks is that they require a significant amount of training data. Given that the medical dataset is small (only 1142 patients). The best results obtained were with a Long Short Term neural network that keeps memory of previous past activities in order to predict the type of cancer of the patients. However, due to the lack amounts of training data, the algorithm could not generalise well and all models found using a grid search approach showed some levels of overfitting as it can be seen in 

\subsection{From Predictability to Explainability using Autoencoders}\label{sec:exp1_interp}

Understanding the reasons why deep learning algorithms make certain predictions, play an important and fundamental role to assess the effectiveness of the model and as well as providing new insights of how to transform a system or a prediction that is untrustworthy to a trustworthy one. 

In this section, we investigate how the different algorithms in Table~\ref{tab:results_exp1.1} are classifying the patients' cancers by using autoencoders. Autoencoders were originally proposed by~\cite{Kramer91} and are unsupervised learning techniques which use neural networks for the task of representation learning. The network architecture enables a compression of knowledge representation of the original input. This implies that correlated features provide a structure that can be learned by the network and consequently one can obtain visualisations of neurons that are being activated in the hidden layer. This compression of knowledge is crucial for the network architecture, since without its presence, the network could simply learn to copy the input values and propagate them throughout the network~\cite{harradon2018causal}. 

The structured deep learning network that was learnt using different layers fuses different modalities of information, based not only on the patients' track of medical activities, but also other features such as age, time spent in treatment, etc. This fusion of information is non-linear and leads to the representation of one single state of knowledge. 

To gain understandings about the network's structured representation of this state of knowledge, we intercepted the first hidden layer of both the LSTM and BiLSTM neural networks in Figure~\ref{fig:deep_arch} and applied an autoencoder to learn the input that led to the projections in this hidden layer. To be more specific, we used an autoencoder with two dense layers to learn the generalized latent space that better approximates to the training data. From the structured deep learning network, the autoenconders apply a non-linear transformation in the data that leads to a non-linear representation of clusters that can be helpful to provide additional insights and that can enable the investigation of misclassifications in the dataset. This provides better insights of why the algorithm is classifying the data correctly or incorrectly, and new understandings to the decision-maker of 

\begin{figure}[h!]
    \resizebox{\columnwidth}{!} {
    \includegraphics{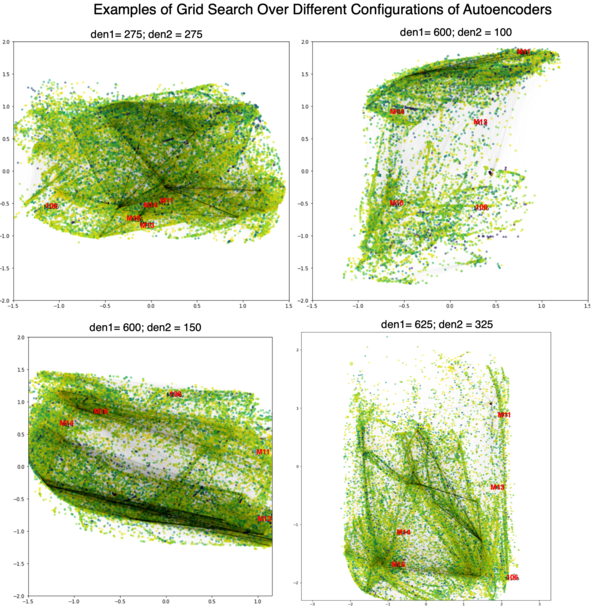}
    }
    \caption{Example of grid searches over the first hidden layer of the LSTM network for different configurations neurons in each of the two dense layers of the autoencoder.}
    \label{fig:example_grid_lstm}
\end{figure}

\begin{figure}[h!]
    \resizebox{\columnwidth}{!} {
    \includegraphics{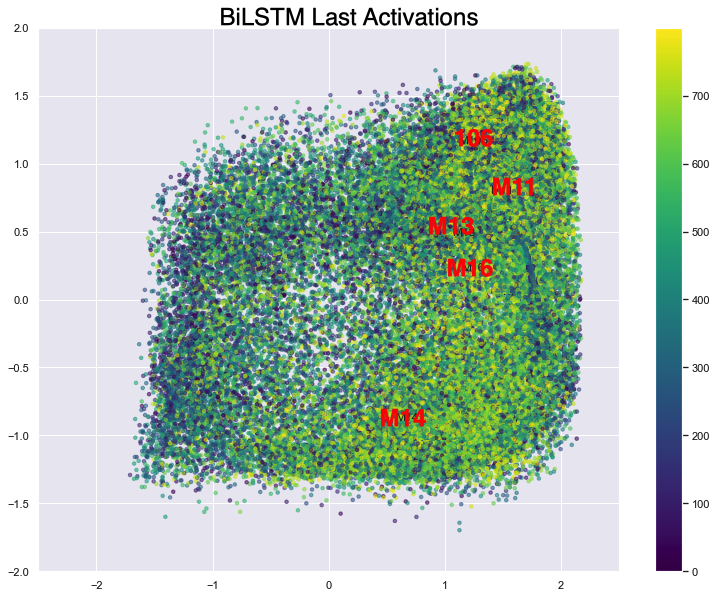}
    }
    \caption{Example of an output image from a grid search approach that intercepts the first BiLSTM layer of the proposed deep neural network architecture. Some clusters identifying the projections of the types of cancer can be found}
    \label{fig:example_grid_bilstm}
\end{figure}

A grid search approach was used in order to find autoencoders that could provide meaningful results to the decision-maker regarding the relationships between the patients features and their types of cancer. Figures~\ref{fig:example_grid_bilstm} and~\ref{fig:example_grid_lstm} show examples of projections that were obtained using an autoencoder with two dense layers and different number of neurons for the BiLSTM and LSTM layers, respectively.

After performing a grid search, we extracted the most meaningful representations from the non-linear projections of the autoencoders, both for LSTM and BiLSTM network architectures, in order to analyse the misclassifications in each model. Figures~\ref{fig:example_grid_lstm} and~\ref{fig:example_grid_bilstm}, show the general latent spaces that were extracted for the LSTM model and BiLSTM model, respectively.
 
 \begin{figure}[h!]
     \resizebox{\columnwidth}{!} {
     \includegraphics{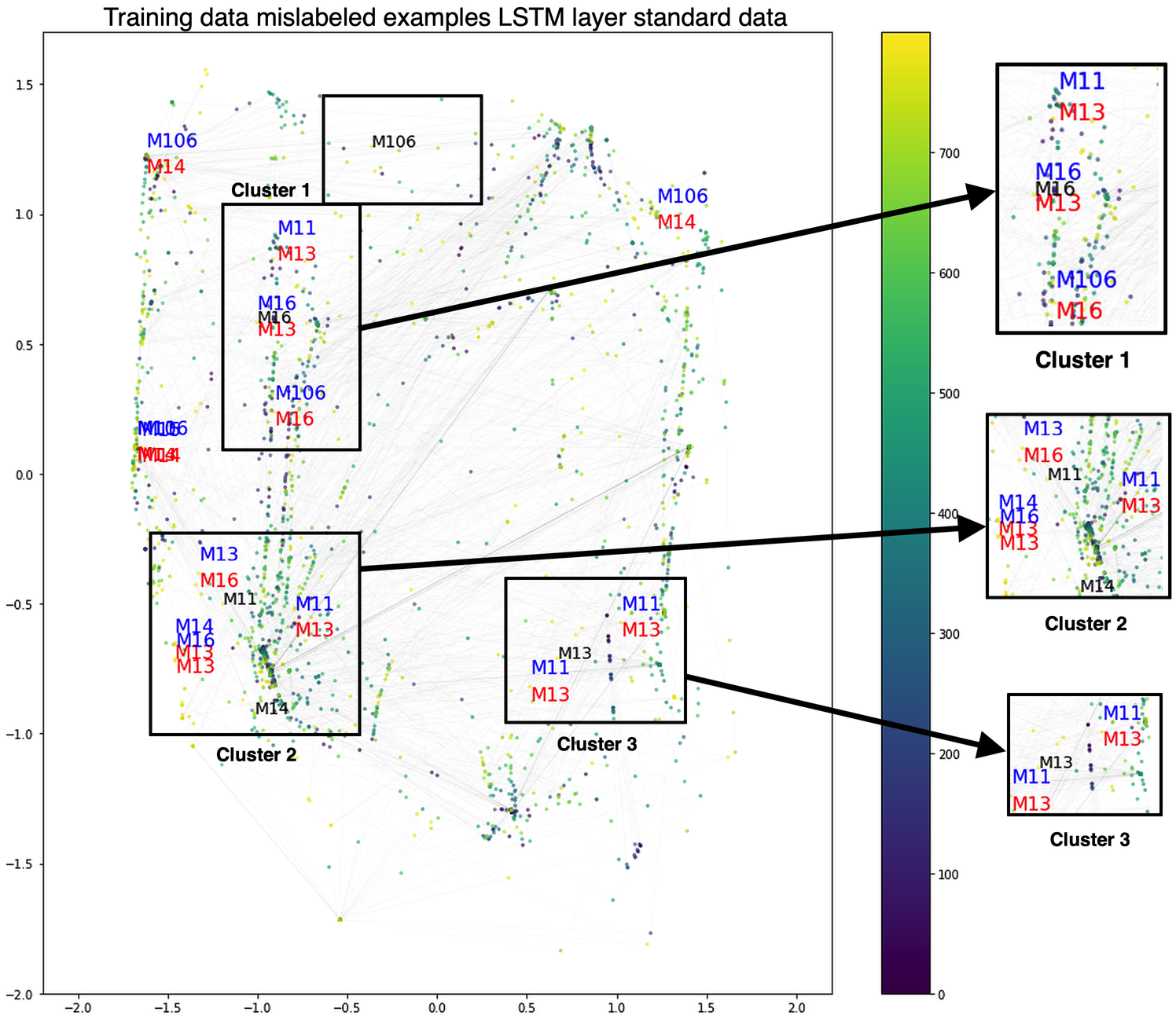}
     }
     \caption{Misclassifications found in the projections of the Long Short Term Memory network with autoencoders.}
     \label{fig:misclassifications_lstm}
 \end{figure}
 
 Sparser results were obtained in the LSTM model, which enabled the identification of non-linear cluster representation in this latent space representation of the state of knowledge of the network. As one can see in Figure~\ref{fig:misclassifications_lstm}, one is able to find three different clusters of data: (1) cluster 1, M16 (cancer of ovary), (2) cluster 2, M11 and M14 (cancer of vulva and cancer of corpus uteri), and (3) cluster 3, M13 (cancer of cervix).

In all three non-linear clusters that were identified, one can see that patients with different types of cancer were projected to the wrong clusters. For instance, in cluster 1, that is identified as the cluster with patients with cancer of ovary (M16), there are patients that have the label cancer of vulva (M11) and yet the model classifies not as M16, but as M13 (cancer of cervix). This can either mean two things, (1) this specific patient diagnosed with M11 shares a very similar track of medical activities as a patient diagnosed with M13 or (2) the non-linearity nature of the projections in the generalised latent representation into a lower dimension space, distorted the distances between these patients, and as a consequence they were assigned to the wrong cluster. 

When it comes to diagnosis code 106, which pertains to patients with a mix of cancers (cervix, vulva, corpus uteri and vagina), projections from the general latent representation into lower dimensions did not show specific misclassifications around this code in that specific region of the lower dimensional space. However, since code 106 pertains with patients with different types of cancer, which according to the figure mainly intersects M14 (cancer of corpus uteri), then one can understand the different misclassifications around patients with code 106 throughout the space.

\begin{figure}[h!]
     \resizebox{\columnwidth}{!} {
     \includegraphics{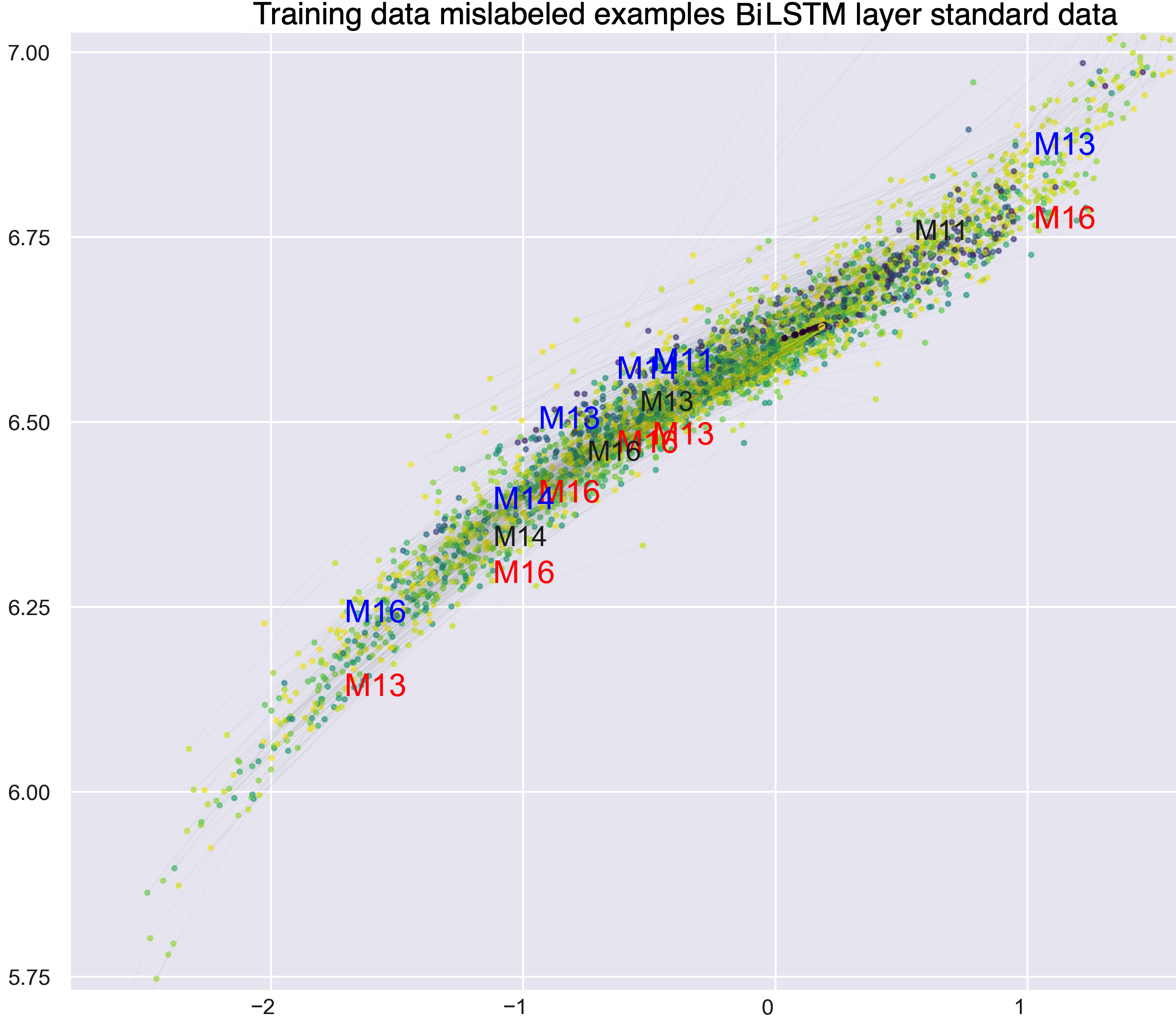}
     }
     \caption{Misclassifications found in the projections of the Bidirectional Long Short Term Memory network with autoencoders.}
     \label{fig:misclassifications_bi}
\end{figure}
 
On the other hand, the non-linear projections found in the BiLSTM (Figure~\ref{fig:misclassifications_bi}) did not show a clear understanding when compared with the projections in the LSTM network. This is due to the fact that during the grid search process, no sparse representations were found, which makes the representation of the space very compact. One can, however, still gain insights about the misclassifications. Like it was found in the LSTM layer, patients with cancer of vulva (M11) were wrongly projected into the M13 cluster (cancer of cervix). Once again, the non-linearity of the projections, together with the mapping into lower dimensions, disturbs the space and the distances between the patients, leading to misclassifications.
 
\section{Experiment II: Explanatory Mechanisms for Predictions Using Random Forests}\label{sec:exp2}

In this section, we explore alternative sub-symbolic representations and understandings of data using random forests and by learning an interpretable model locally around the model's predictions.

\subsection{Problem Definition}\label{sec:exp2_problem}
The problem is converted to a classical supervised learning problem to compare and contrast traditional approaches while using event logs to predict cancer. Here, for each patient $X^{(i)}$, the set of features $F$ (both dynamic and static) are mapped to the window of length $T$. The window represents the daily (or hourly) medical activities associated to a patient. A patient $X^{(i)}$ is represented by the vector: 
$\langle f_{1,1}^{(i)}, f_{1,2}^{(i)}, \hdots, f_{1,F}^{(i)} \hdots  f_{T,1}^{(i)}, f_{T,2}^{(i)}, \hdots, f_{T,F}^{(i)} \rangle$. Hence, $M$ patients are represented by a matrix with dimensions $(M \times  (F*L) )$. The length $L$ is the number of patient's medical records (or activities recorded for each patient). The cancer associated to each patient is the class we predict. The advantage of this approach is that it allows any classical supervised machine learning algorithm to be applied.

\subsection{Random Forests for Cancer Prediction in Event Logs}\label{sec:exp2_algo}

Random forests are an ensemble method that combine several individual classification trees \cite{Breiman2001}. A Random forest classifier uses multiple decision tree classifiers where each decision tree classifier is fit to a random sample, or a bootstrap sample drawn from the original data sample. The feature selected for each split in the classification tree is only from a small random subset of features in each tree. Thus, a random forest classifier consists of a number of classification trees, the value of which is set when identifying the model parameters.  From the forest, the class or label is predicted as an average or majority vote of the predictions of all trees. 

Random forests are known to have high prediction accuracy as compared to individual classification trees, because the ensemble adjusts for the over-fitting caused by individual trees. However, the interpretability of a random forest is not as straightforward as that of an individual tree classifier, where the influence of a feature variable corresponds to its position in the tree.

\subsection{Predicting Patient's Type of Cancer}\label{sec:exp2_res}
\begin{figure}[h!]
	 \resizebox{\columnwidth}{!} {
	 \includegraphics{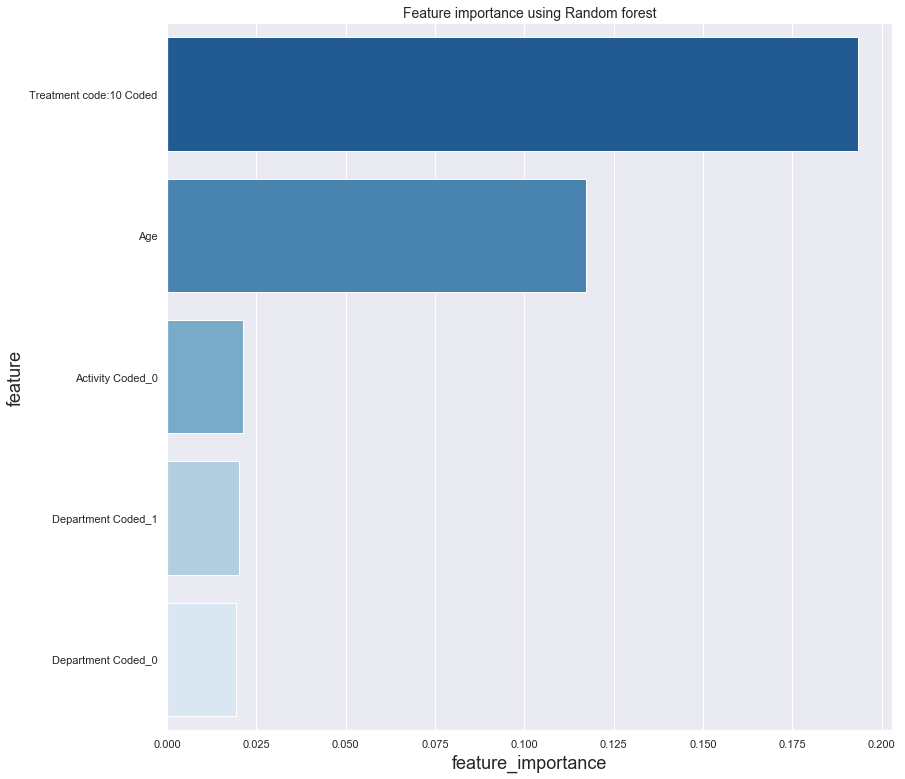}
	 }
	  \caption{Top 5 important features used by Random Forest classifier.}
	   \label{fig:feature_imp_rf}	
\end{figure}

To validate the Random forest classifier, we performed a cross validation setting with a train/test set split of $80\% / 20\%$. The optimal parameters for the classifier were found using grid search with k-fold cross validation. Table~\ref{tab:results_exp_rf} presents the accuracy for two different parameters used during the grid search parameter tuning. Figure~\ref{fig:feature_imp_rf} presents the top five important predictors. This plot shows the features such as the $Age$ of the patient, the type of $Treatment$, and initial set of $Activity$ performed in a given sequence during the treatment ($Activity~Coded\_0$, $Activity~Coded\_1$, $Activity~Coded\_2$ representing $Activity\_(sequence\_number)$ are among the most important features for predicting the cancer.

\begin{table}[h!]
    \centering
    \begin{tabular}{l |c|c |}    
     {\bf Estimators}   & {\bf Maximum features}    & {\bf Accuracy}    \\
        \hline
{ 1000}           & 100       & 0.556     \\
{ 1500}           & 200  & 0.572  \\
           \hline
    \end{tabular}
    \caption{Results obtained while conducting a cross validation grid search over the the number of estimators and size of the random subsets of features used for splitting a node in the tree.}
    \label{tab:results_exp_rf}
\end{table}

The importance of a feature when using a Random forest classifier is computed using the `gini impurity' measure that indicates the effectiveness of a feature in reducing uncertainty when creating decision trees. However, this method tends to inflate the importance of continuous or high-cardinality categorical variables \cite{Strobl2007}. Hence, while feature importance using `gini impurity' measure has been consistently used, it is provides interpretability of the entire model and does not provide explanation of a specific instance.

\subsection{From Predictability to Explanability using LIME}\label{sec:exp2_interpr}
LIME \cite{Ribeiro16} is used to explain a single prediction as well a global explanation of the model using a subset of individual data points or instances. LIME approximates the underlying model by an interpretable model such as a linear model that is learned on small perturbations of the original data point. This is done by weighting the perturbed instance by their similarity to the instance to be explained. Hence, the explanations are based on a linear model in the neighborhood of the instance and the explanations for an instance does not represent how the model behaves for all data points or cancer patients.
\begin{figure}[h!]
	 \resizebox{\columnwidth}{!} {
	 \includegraphics{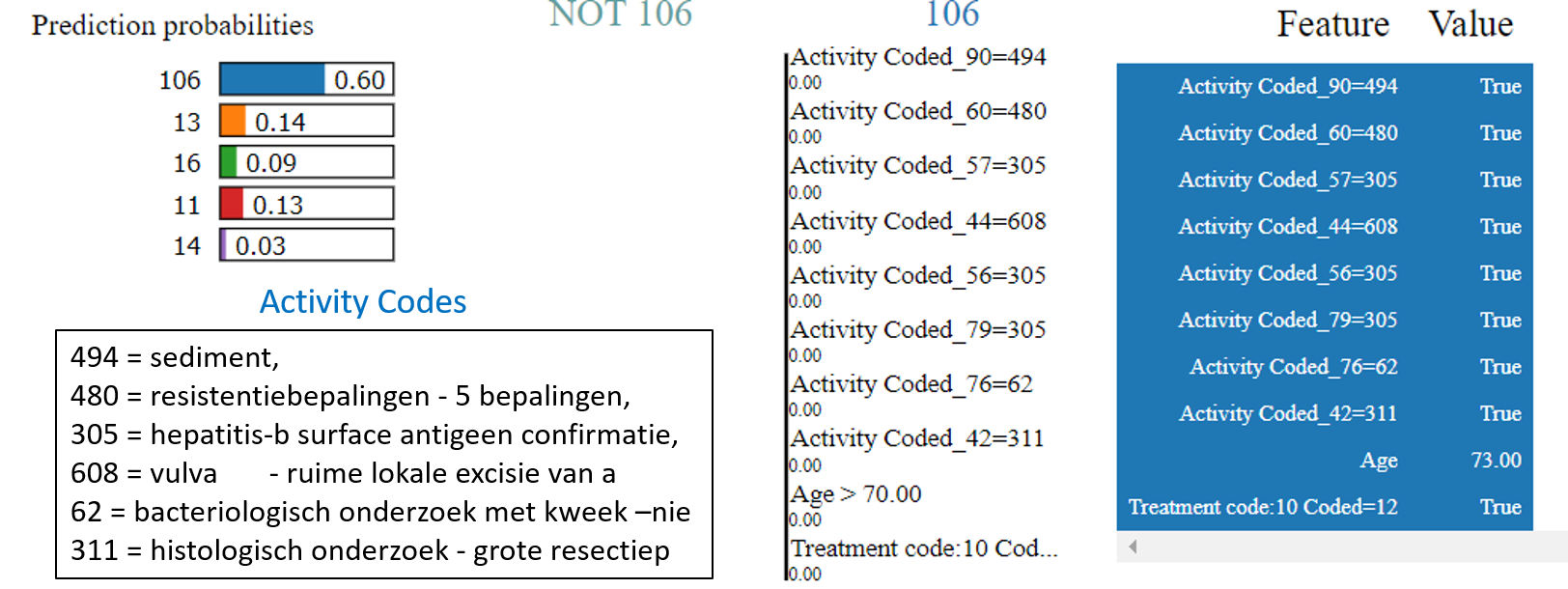}
	 }
	  \caption{Local interpretation of 106 cancer class for a patient.}
	   \label{fig:lime_local_106}	
\end{figure}
Figure~\ref{fig:lime_local_106} illustrates the local explanations of predicting the cancer class `106' which is associated to cervix, vulva, corpus uteri and vagina. The explanations are based on the features \textit{Age $>$ 70}, and specific activities performed at a given step or sequence during the treatment ( \texttt{Activity\_(sequence\_number)} ).

\begin{figure}[h!]
	 \resizebox{\columnwidth}{!} {
	 \includegraphics{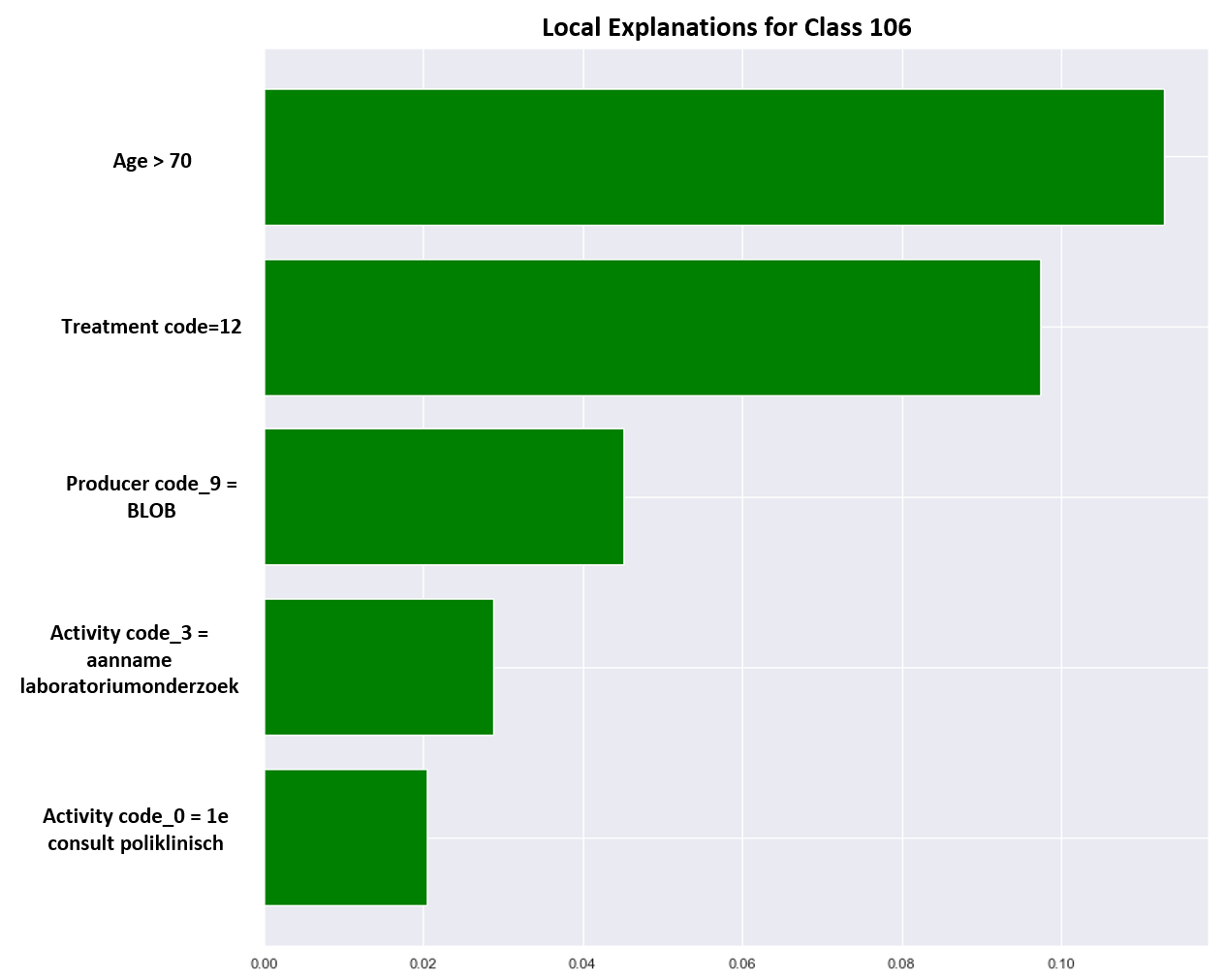}
	 }
	  \caption{Global interpretation of class 106 cancer.}
	   \label{fig:lime_global_106}	
\end{figure}

\begin{figure}[h!]
	 \resizebox{\columnwidth}{!} {
	 \includegraphics{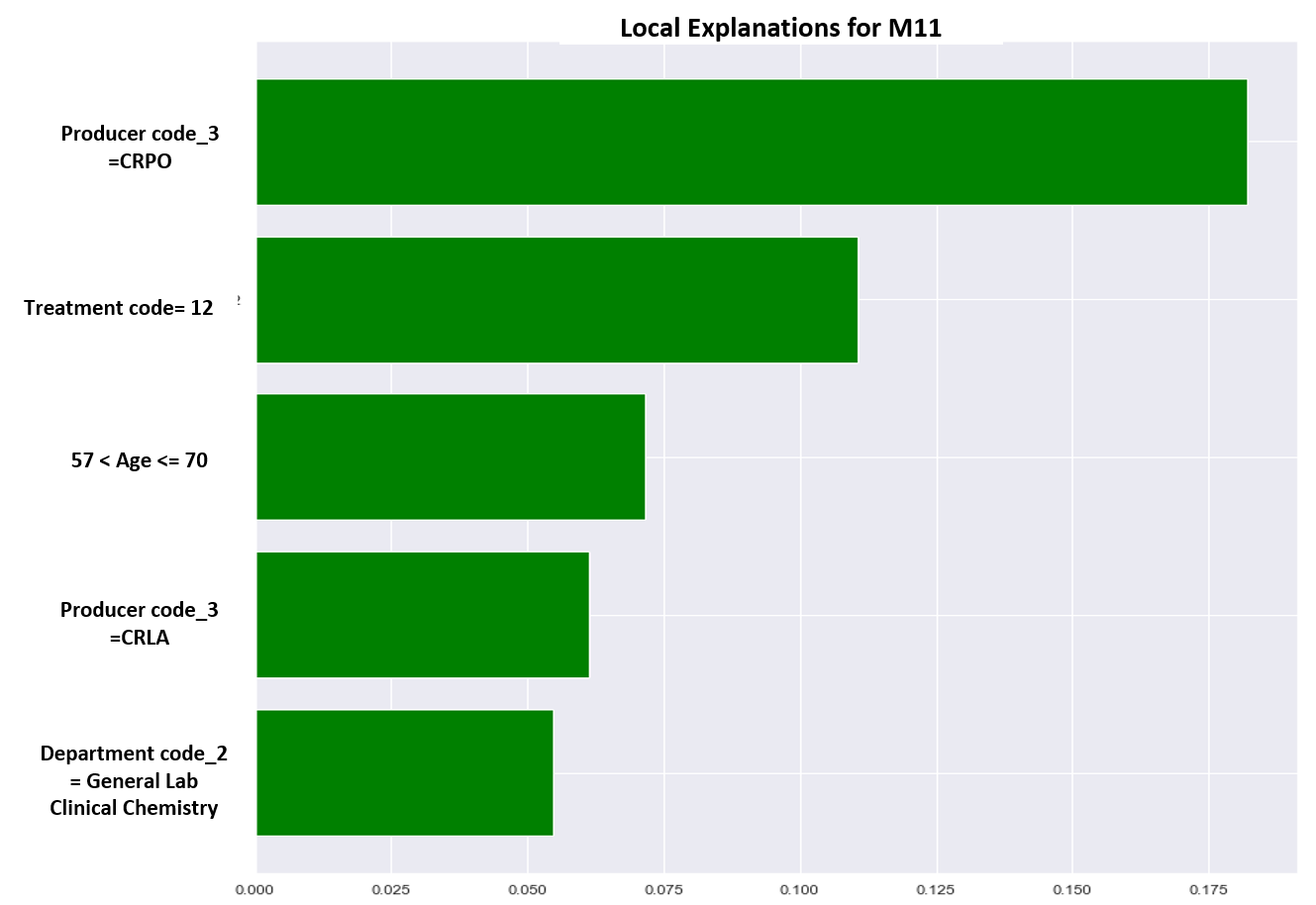}
	 }
	  \caption{Global interpretation of class M11 cancer.}
	   \label{fig:lime_global_11}	
\end{figure}

\begin{figure}[h!]
	 \resizebox{\columnwidth}{!} {
	 \includegraphics{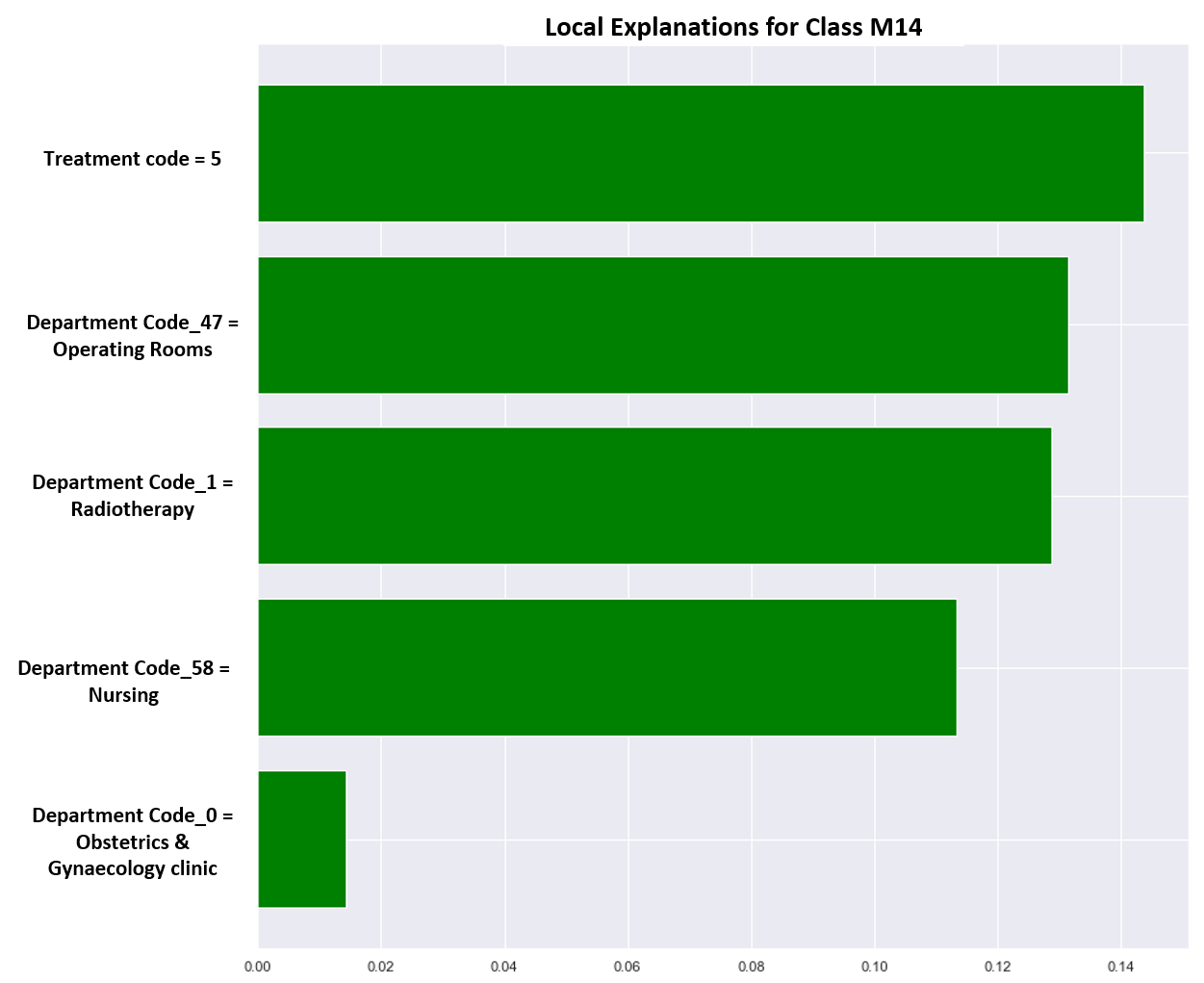}
	 }
	  \caption{Global interpretation of class M14 cancer.}
	   \label{fig:lime_global_14}	
\end{figure}

The global understanding of the model is provided by explaining a set of individual instances. The global explanations of the model are constructed by picking a subset of instances and their explanations. The importance of a feature in an explanation and the coverage of all features defines a coverage function that is maximized to pick  a subset of instances and generate global explanations. Figure~\ref{fig:lime_global_106} presents the global explanation for the cancer class `106'. Here the age, the treatment and activities performed initially provide explanation of the predictions. 
Global explanations for two cancer classes (M11, M14) are presented in Figure~\ref{fig:lime_global_11} and Figure~\ref{fig:lime_global_14} respectively. While some of the features used by the model are relevant such as Age and the treatment undertaken, many features such as the activity `Consultation', or being associated to the `Obstetrics \& Gynaecology clinic' are not significantly distinct features and cannot be generalized in predicting the type of cancer. 
However, use of such explanations provides good insight into the model and improves the trust in the prediction, and the features used for the prediction. In the context of traditional machine learning algorithms, use of local explanations provide insights on the design of features.

\section{Conclusions}\label{sec:conclusions}

In this work, we explored the usage of deep learning techniques and random forests in a real world medical event log from a hospital in the Netherlands containing the track of medical records undertaken by patients with cancer. Our hypothesis was that, in theory, patients with a specific type of cancer should be associated to a more targeted set of medical activities that are particular to their type of cancer. Results showed significant results and that one could actually predict the type of cancer given past medical records of patients. The structured learning models that we explored learnt to fuse different modalities of information, based not only on the patients' track of medical activities, but also other features such as age, time spent in treatment, etc. This fusion of information is non-linear and leads to the representation of one single non-linear state of knowledge. However, this analysis in terms of accuracies can be misleading since we do not have any understandings of how the learning algorithms were making the classification.

In this sense, this paper also explored explanability and interpretability techniques in the scope of medical event logs. In order to gain more insights about the model's black box, we intercepted the hidden layers of deep neural networks with autoencoders in order to learn a generalized latent space that better approximates to the training data. From the structured deep learning network, the autoenconders apply a non-linear transformation in the data that leads to a non-linear representation of clusters that can be helpful to provide additional insights and that can enable the investigation of misclassifications in the dataset. This  method provided better insights of why the algorithm is classifying the data correctly or incorrectly, and provided new understandings to the decision-maker.

For random forests, we explored local surrogate models, more specifically the local interpretable model-agnostic explanations (LIME) framework. LIME is a metamodel that instead of interpreting directly the black box, it uses the metamodel to draw conclusions and interpretations about the black box. The individual predictions were computed by applying perturbations of the points in the original dataset. This allows one to see how the features change around these points and how they affect the predictions. Results indicate that learning an interpretable model locally around the model's prediction leads to a higher understanding about why the algorithm is making some decision. The use of local and linear model helped to identify the features used during the cancer prediction of an individual patient. We were able to identify distinct features used in different predictions, along with features that do not generalize or are not relevant.

In summary, both methods provided different sub-symbolic interpretation insights, one based on non-linear cluster representations (autoecoders) and the other based on the local impact of features in individual points in the data (LIME).

\section{Acknowledgements}

Dr. Andreas Wichert was supported by funds through Funda\c{c}\~{a}o para a Ci\^{e}ncia e Tecnologia (FCT) with reference UID/CEC/50021/2019. The funders had no role in study design, data collection and analysis, decision to publish, or preparation of the manuscript.


\bibliographystyle{IEEEtran}


\end{document}